\documentclass{ecai} 

\usepackage{latexsym}
\usepackage{amssymb}
\usepackage{amsmath}
\usepackage{amsthm}
\usepackage{booktabs}
\usepackage{enumitem}
\usepackage{graphicx}
\usepackage{color}
\usepackage{hyperref}

\usepackage{subfig}
\newtheorem{prop}{Proposition}

\usepackage{amsfonts}
\usepackage{algorithm}

\usepackage{algpseudocode}

\usepackage{xurl}

\usepackage{listings}

\usepackage{tabularx}

\usepackage{textcomp}

\usepackage{xcolor}

\usepackage{comment}

\usepackage{stmaryrd}
\usepackage{dsfont}

\usepackage{multirow}
\usepackage{bm}
\newcolumntype{C}[1]{>{\centering\arraybackslash}p{#1}}

\newcommand{\BibTeX}{B\kern-.05em{\sc i\kern-.025em b}\kern-.08em\TeX}

\begin{document}


\begin{frontmatter}


\paperid{123} 


\title{Enhancing Fairness through Reweighting: A
Path to Attain the Sufficiency Rule}


\author[A]{\fnms{Xuan}~\snm{Zhao}\thanks{Corresponding Author. Email: zhaoxuan00707@gmail.com}}
\author[A]{\fnms{Klaus}~\snm{Broelemann}}
\author[B]{\fnms{Salvatore}~\snm{Ruggieri}}
\author[C]{\fnms{Gjergji}~\snm{Kasneci}}

\address[A]{SCHUFA Holding AG}
\address[B]{University of Pisa}
\address[C]{Technical University of Munich}


\begin{abstract}
We introduce an innovative approach to enhancing the empirical risk minimization (ERM) process in model training through a refined reweighting scheme of the training data to enhance fairness. This scheme aims to uphold the sufficiency rule in fairness by ensuring that optimal predictors maintain consistency across diverse sub-groups. We employ a bilevel formulation to address this challenge, wherein we explore sample reweighting strategies. Unlike conventional methods that hinge on model size, our formulation bases generalization complexity on the space of sample weights. We discretize the weights to improve training speed. Empirical validation of our method showcases its effectiveness and robustness, revealing a consistent improvement in the balance between prediction performance and fairness metrics across various experiments. Code is available at \url{https://github.com/zhaoxuan00707/Reweighting_for_sufficiency}.
\end{abstract}

\end{frontmatter}


\section{Introduction}\label{sec:intro}

Machine learning has found extensive application in real-world decision-making processes, including areas such as health care systems \cite{DBLP:conf/kdd/AhmadPEKT20}. 
Algorithmic fairness has garnered significant attention as a means to mitigate predictive bias linked to protected features such as ethnicity, gender, or age. Consequently, numerous fairness notions catering to diverse objectives have been proposed. While many existing approaches in classification or regression adhere to independence or separation rules (refer to Section \ref{sec:related} and related references) \cite{DBLP:conf/icml/MadrasCPZ18,DBLP:conf/aistats/SongKGZE19,DBLP:conf/nips/ChzhenDHOP20}, it's worth noting that these rules may not always be suitable in various applications. In such cases, alternative fairness notions, such as the sufficiency rule \cite{DBLP:journals/bigdata/Chouldechova17}, are favored. In simple terms, the sufficiency rule, detailed in Section \ref{sec:related}, ensures that the conditional expectation of $\mathbb{E}[Y | \hat{Y}]$ remains consistent across different sub-groups, providing a more nuanced approach to fairness.

In practical terms, neglecting the sufficiency rule can result in significant biases within intelligent healthcare systems. For instance, many health systems utilize algorithms to identify and support patients with complex health requirements. These algorithms generate a score indicating the level of healthcare needs, with higher scores suggesting greater sickness and the need for more care. Notably, a study by \cite{Obermeyer2019} uncovers a widely used industry algorithm affecting millions of patients, which exhibits pronounced racial bias. It was found that for a given predicted score $\hat{Y}=t$, black patients tend to be considerably sicker than white patients ($\mathbb{E}_{black}[Y | \hat{Y} = t] > \mathbb{E}_{white}[Y | \hat{Y} = t]$). Moreover, the study highlights that rectifying this disparity could significantly increase the percentage of black patients receiving additional care from 17.7\% to 46.5\%. From an algorithmic perspective, the sufficiency rule is generally incompatible with concepts such as independence or separation, as demonstrated in Section \ref{sec:related} and the Appendix \cite{zhao2024ecai}. This suggests that existing fair algorithms designed for independence or separation may not enhance or could even exacerbate issues related to the sufficiency rule. Notably, recent work on Invariant Risk Minimization (IRM) proposed by \cite{DBLP:journals/corr/abs-1907-02893,bühlmann2018invariance} has potential to address this challenge. IRM seeks to maintain invariant correlations between the embedding (or representation) and the true label by incorporating regularization techniques into Deep Neural Network (DNN) training. The criteria of the sufficiency rule and IRM are intrinsically consistent (see more details in Section \ref{consistent}); the idea being that if the correlations between the embedding (or representation) and the true label remain robust and unaffected by specific sub-groups, the resulting representation can be considered fair. 
To better understand this concept, IRM addresses the challenge of ensuring that \textit{a cow is correctly classified as a cow in a picture, regardless of whether the background is Grass or Desert}~\cite{DBLP:journals/corr/abs-1907-02893}. On the other hand, the sufficiency rule aims to ensure that \textit{a patient predicted to be high-risk is truly high-risk, regardless of whether the patient is Black or White}. IRM approaches have attracted attention due to their promising performance on modest models and datasets~\cite{DBLP:journals/corr/abs-1907-02893} and their simplicity in facilitating end-to-end training. Nevertheless, recent studies have indicated diminished effectiveness of the regularization terms when applied to overparameterized DNNs~\cite{DBLP:journals/corr/abs-2102-06764,DBLP:conf/cvpr/LinDWZ22}. For instance, CelebA comprises only 200k training data, whereas ResNet-18 boasts 11.4 million parameters. Overparameterized DNNs can easily diminish the regularization term of IRM to zero during training while still depending on spurious features. In such scenarios, applying IRM methods directly for fairness to uphold the sufficiency rule in relatively larger models is deemed inappropriate. 

This paper introduces a novel approach to address the aforementioned limitation by proposing a model-agnostic sample reweighting method. Our method 
transforms the parameter search space of the model into one of sample weights by formalizing the learning of sample reweighting as a bilevel optimization problem. Within the \textbf{inner loop}, we train DNN on the weighted training samples. In the \textbf{outer loop}, we employ the IRM criterion as the outer objective to guide the learning process of the sample weights, thereby enforcing the sufficiency rule. We iteratively alternate between the inner and outer loops, ultimately obtaining a set of weights~$w$ with an advantageous characteristic: utilizing only learned sample weights on training samples, we can conduct weighted empirical risk minimization (ERM) training to achieve superior fairness.

Our contributions are summarized as follows:
\begin{enumerate} 
\item We introduce a model-agnostic sample reweighting approach rooted in bilevel optimization for IRM learning to promote fairness. This method offers notable advantages, particularly in transforming the optimization problem from the parameter space of DNNs to the space of sample weights. This shift effectively mitigates the overfitting issues commonly encountered by IRM regulariza\-tion-based methods.

\item 
Our method formulate the fairness issue as a bilevel optimization and does not impose specific fairness constraints, thus avoiding the issue of determining critical hyperparameters for fairness regularization. 

\item We substantiate the superior performance of our approach through empirical evaluations across diverse tasks, showcasing its effectiveness compared to state-of-the-art methods.

\end{enumerate}

The structure of this paper is as follows. In Section \ref{sec:related}, we present a comprehensive review of the notation and background related to fairness notions, IRM, and reweighting methods. Section \ref{sec:method} outlines our sample reweighting method in detail. In Section \ref{sec:exe}, we conduct experiments to compare the accuracy and fairness across four datasets against state-of-the-art methods, illustrating the robustness and effectiveness of our framework.

\section{Preliminaries and Related Work} \label{sec:related}
\subsection{Sufficiency Rule in Fairness}
We denote the predictive features as $X \in \mathcal{X}$, the ground truth label as $Y \in \mathcal{Y}$, and the algorithm's output as $\hat{Y} \in \mathcal{Y}$. We consider a binary protected feature or two sub-groups $\mathcal{D}_0$ and $\mathcal{D}_1$. Then, in accordance with \cite{DBLP:conf/icml/LiuSH19}, the sufficiency rule is defined as follows:

\begin{equation}\label{eq:suf}
\mathbb{E}_{\mathcal{D}_0} [Y | \hat{Y} = t] =\mathbb{E}_{\mathcal{D}_1} [Y | \hat{Y} = t], \forall t \in \mathcal{Y}
\end{equation}

\subsubsection{Sufficiency Gap}
Eq. (\ref{eq:suf}) indicates that the conditional expectation of the ground truth label $Y$ is consistent across both $\mathcal{D}_0$ and $\mathcal{D}_1$, given the same prediction output $s$. In \cite{DBLP:conf/icml/ShuiC0W022}, the sufficiency gap is proposed as a metric for fairness measurement. In binary classification, the sufficiency gap is naturally defined as follows:

\begin{equation}
\Delta\text{Suf}=\frac{1}{2}\sum\limits_{y \in \{0,1\}}|P_{\mathcal{D}_0}(Y=y | \hat{Y} = y)-P_{\mathcal{D}_1}(Y=y | \hat{Y} = y)|
\end{equation}

The sufficiency gap $\Delta\text{Suf} \in [0, 1]$. A value close to $0$ indicates equality between two sub-groups, which have close Positive Predictive Values (PPV) and Negative Predictive Values (NPV). To grasp the significance of this metric, consider a healthcare system that only outputs binary scores: High Risk or Low Risk. As highlighted in \cite{Obermeyer2019}, if $P_{\mathcal{D}_{\text{black}}}(Y = \text{High Risk}| \hat{Y} = \text{Low Risk}) \gg P_{\mathcal{D}_{\text{white}}}(Y = \text{High Risk}| \hat{Y} = \text{Low Risk})$, then the severity of illness is underestimated more for black patients than for white patients. Therefore, a small value of $\Delta\text{Suf}$ indicates that racial discrimination is addressed.

\subsubsection{Relation to Other Fairness Notions}



We briefly contrast the Sufficiency rule with the commonly employed Independence and Separation rules in binary classification. For comprehensive justifications and comparisons, please consult Appendix \cite{zhao2024ecai}.

The \emph{Independence rule} is:

\begin{equation}
\mathbb{E}_{\mathcal{D}_0} [ \hat{Y} ] = \mathbb{E}_{\mathcal{D}_1} [\hat{Y}]
\end{equation}
In binary classification, the Independence rule is often referred to as demographic parity (DP) \cite{DBLP:conf/icml/ZemelWSPD13}. Furthermore, it can be argued that if $P_{\mathcal{D}_0}(Y = y) \neq P_{\mathcal{D}_1}(Y = y)$ (indicating distinct label distributions in the sub-groups), it is impossible for both the Sufficiency and Independence rules to hold simultaneously \cite{Castelnovo_2022}.

\emph{Separation Rule} is:
\begin{equation}
\mathbb{E}_{\mathcal{D}_0} [ \hat{Y} |Y = t] = \mathbb{E}_{\mathcal{D}_1} [ \hat{Y} |Y = t], \forall t \in Y
\end{equation}

In binary classification, the Separation rule is also referred to as Equalized Odds (EO) \cite{DBLP:conf/nips/HardtPNS16}. Additionally, \cite{barocas-hardt-narayanan} have further illustrated that if $P_{\mathcal{D}_0}(Y = y) \neq P_{\mathcal{D}_1}(Y = y)$ and the joint distribution of $(Y,\hat{Y})$ has a positive probability in $\mathcal{D}_0$ and $\mathcal{D}_1$, then it is impossible for both the Sufficiency and Separation rules to coexist \cite{Castelnovo_2022} (please refer to Appendix\cite{zhao2024ecai} for further details).

\subsection{Invariant Risk Minimization} \label{sec:IRM}


IRM operates under the assumption that there are multiple environments $\mathcal{E} :=\{e_1, e_2,..., e_E\}$ within the sample space $\mathcal{X}\times \mathcal{Y}$, each characterized by distinct joint distributions. Furthermore, it assumes that the correlation between the spurious features and labels varies inconsistently across these environments. The predictor $f (\cdot; \theta)$ in IRM is expressed as a composite function of a representation $\phi(\cdot; \Phi)$ and a classifier $h(\cdot; v)$, formulated as $f (\cdot;\theta) = h(\phi(\cdot; \Phi); v)$, where $\theta= \{v, \Phi\}$ represents the trainable parameters. The fundamental idea is that if a predictor $f (\cdot; \theta)$ performs effectively across all environments, it suggests that the correlation between the spurious features and labels is not accurately captured\cite{peters2015causal,DBLP:journals/corr/abs-1907-02893}. In these cases, the data representation function $\phi$ elicits an invariant predictor across environments $\mathcal{E}$ if and only if for all latent $z$ in the intersection of the supports of $\phi(X^e)$ we have $\mathbb{E}[Y^e|\phi(X^e) = z] = \mathbb{E}[Y^{e'}|\phi(X^{e'}) = z]$, for all $e, e' \in \mathcal{E}$ (For loss functions such as the mean squared error and
the cross-entropy, optimal classifiers can be written as conditional expectations). Please refer to Appendix\cite{zhao2024ecai} for more details. Consequently, IRM aims to minimize a specific IRM risk to identify such a robust predictor. Several approaches have been proposed to enhance IRM: \cite{DBLP:conf/icml/KruegerCJ0BZPC21,DBLP:journals/corr/abs-2006-07544} advocate for penalizing the variance of risks across different environments, while \cite{DBLP:conf/icml/ChangZYJ20,DBLP:journals/corr/abs-2110-09940} attempt to estimate the violation of invariance by training neural networks. Moreover, theoretical guarantees for IRM on linear models with adequate training environments are provided by \cite{DBLP:journals/corr/abs-1907-02893,DBLP:conf/iclr/RosenfeldRR21,DBLP:conf/nips/ChenRS0R22}.


Two popular risks are:

\begin{eqnarray} \label{IRMv1}
\mathcal{R^{\text{IRMv1}}}(\mathcal{D},\theta) :=\sum\limits_{e}\mathcal{L}(\mathcal{D}^e,\theta)+\lambda\|\nabla_v\mathcal{L}(\mathcal{D}^e,\theta)\|^2_2
\\
\mathcal{R^{\text{REx}}}(\mathcal{D},\theta) :=\sum\limits_{e}\mathcal{L}(\mathcal{D}^e,\theta)+\lambda\mathbb{V}_e[\mathcal{L}(\mathcal{D}^e,\theta)]
\end{eqnarray}


where $\mathcal{D} = \bigcup^e\mathcal{D}^e$ denotes the data drawn from all environments, where $\mathcal{D}^e$ represents the data from environment $e$. The expression $\mathbb{V}_e[\mathcal{L}(\mathcal{D}^e, \theta)]$ signifies the variance of the loss across various environments.

However, it has been observed that IRM exhibits diminished efficacy when applied to overparameterized neural networks \cite{DBLP:conf/iclr/GulrajaniL21,choe2020empirical}. \cite{DBLP:conf/cvpr/LinDWZ22} elucidates that this limitation can largely be attributed to the problem of overfitting. Consequently, utilizing these methods directly for addressing fairness concerns is not straightforward.


\subsection{Reweighting} Sample reweighting constitutes a classical approach for addressing various tasks such as distribution shifts, imbalanced classification, and fairness concerns. Here, we specifically delve into reweighting methodologies associated with fairness considerations. Fairness with Adaptive Weights \cite{chai2022} imposes constraints on the sum of weights across sensitive groups to ensure equality, assigning weights to each sample based on its likelihood of misclassification. Adaptive Sensitive Reweighting to Mitigate Bias \cite{DBLP:conf/www/KrasanakisXPK18} assigns weights to samples based on their alignment with the unobserved true labeling. \cite{DBLP:conf/icml/LiL22a} intricately models the impact of each training sample on fairness-related metrics and predictive utility. Additionally, \cite{DBLP:conf/bigdataconf/ZhaoBRK23} utilizes Neural Networks to reweigh samples, aiming to achieve causal fairness. To the best of our knowledge, no reweighting method has been specifically applied to achieve the sufficiency rule in fairness. Furthermore, our method stands apart from heuristic reweighting methods, as it does not necessitate complex hyper-parameter selection processes.

\section{Reweighting to Achieve Sufficiency Rule}\label{sec:method}



\subsection{Bilevel Formulation of Reweighting}


Given a dataset $\mathcal{D}$ constituted as a set $\{(x_i,y_i)\}^n_{i=1}$, where each $(x_i, y_i)$ is drawn from $\mathcal{X}\times\mathcal{Y}$, the weighted empirical loss is defined as $\mathcal{L}(\mathcal{D}, \theta; w):= \frac{1}{n}\sum_{i=1}^{n}w_il(f(x_i;\theta),y_i)$, with $f (\cdot; \theta)$ representing a neural network parameterized by $\theta$, $l(\cdot,\cdot)$ indicating the loss function (e.g., cross-entropy or least squares loss), and $w_i\in \mathbb{R}^+$ denoting the non-negative weight assigned to each sample.
We formulate the objective of learning sample weights to mitigate reliance on sensitive features as the subsequent bilevel optimization problem:

\begin{eqnarray} \label{eq:ori}
\underset{w\in\mathcal{W}}{\text{min}}\;\mathcal{R}(\mathcal{D},\theta^*(w)),\\
\text{s.t. } \theta^*(w) \in \underset{\theta}{\text{arg min}}\;\mathcal{L}(\mathcal{D},\theta; w) \notag
\end{eqnarray}


Here, $w$ denotes a vector of sample weights with a length of $n$, indicating the importance of each training sample, where each component $w_i$ of $w$ satisfies $w_i \geq 0$. Any IRM Risk $\mathcal{R}(\mathcal{D}, \theta)$ discussed in Section \ref{sec:related} can function as the outer objective. In our subsequent experiments, we employ the risk (\ref{IRMv1}), denoted as IRMv1. Within the inner loop, we minimize the weighted ERM loss on the training samples to derive a model $\theta^*(w)$, while within the outer loop, we evaluate the learned model's reliance on sensitive features through IRM Risk and adjust the sample weights accordingly. By iteratively alternating between the inner and outer loops, the sample weights gradually adjust to a state where they can yield satisfactory IRM/fairness performance via straightforward ERM training. It's worth noting that, instead of different environmental settings as in the IRM scenario, the fairness problem involves distinct sensitive groups, such as $\mathcal{D}_0$ and $\mathcal{D}_1$, as depicted in Section \ref{sec:related}. Although we showcase our approach within the context of binary sensitive groups in this section, it can be readily extended to scenarios involving multi-categorical sensitive groups (refer to the experimental details on the toxic comments dataset and COMPAS dataset in Section \ref{sec:exe}).


Our approach provides the following benefits: 
1) by establishing an implicit mapping from the sample weight space to the model parameter space in the outer loop, where the former consistently remains smaller than the latter in deep learning tasks (as detailed in Section \ref{sec:intro}), we effectively address overfitting issues typically associated with IRM regularization-based methods (the objective of the outer loop); 2) our approach avoids the need to impose specific fairness constraints, thereby circumventing the challenge of determining critical hyperparameters for fairness regularization to achieve a better trade-off between fairness and accuracy. 

\subsubsection{Connection to the Sufficiency Rule:}\label{consistent} 



We elaborate the connection between the outer loop of our bilevel objective and the Sufficiency Rule \cite{DBLP:conf/icml/ShuiC0W022}.

\begin{prop} \label{prop:link}
In a classification task, 
minimizing the loss in the outer loop as illustrated in details in Section \ref{sec:IRM} is tantamount to:

\begin{eqnarray} 
\mathbb{E}_{\mathcal{D}_0} [ Y|Z=z]= \mathbb{E}_{\mathcal{D}_1} [ Y|Z=z],\
\text{(IRM definition)}\\
\mathbb{E}_{\mathcal{D}_0} [ Y |\hat{Y}=h^*(z)] = \mathbb{E}_{\mathcal{D}_1}[Y|\hat{Y}= h^*(z)]
\end{eqnarray} 

where $h^*_0$, $h^*_1$ are the optimal predictor for each sub-group, $h^*=h^*_1=h^*_0$ and $z=\phi(x)$.

\end{prop}

Proposition \ref{prop:link} illustrates that the objective of the outer loop loss aligns with the sufficiency rule in binary classification. 


\subsection{Enhance Reweighting by Sparsity and Continuation}

 We discretize the optimization method \cite{DBLP:conf/icml/ZhouPZLCZ22} here, 

\begin{eqnarray} \label{eq:discrete}
\underset{m\in\mathcal{C}}{\text{min}}\;\mathcal{R}(\mathcal{D},\theta^*(m)),\\
\text{s.t. } \theta^*(m) \in \underset{\theta}{\text{arg min}}\;\mathcal{L}(\mathcal{D},\theta; m) \notag
\end{eqnarray}


where the mask $m \in \{0, 1\}_n$ represents a binary vector, and $m_i=1$ denotes that sample~$i$ is included in the training set, otherwise it is excluded. $K$ is a positive integer that determines the size of the selected set, and $\mathcal{C} = \{m : m_i \in \{0, 1\}, \|m\|_0 \leq K\}$ denotes the feasible region of $m$. Essentially, the inner loop trains the network to converge on the selected set to obtain the model $\theta^*(m)$, while the outer loop assesses the loss of $\theta^*(m)$ on the entire set and optimizes it to guide the learning of $m$.


The distinction between our discrete bilevel formulation (\ref{eq:discrete}) and the original bilevel formulation (\ref{eq:ori}) lies in the absence of individual weights $w_i$ for each sample in the sparse formulation (\ref{eq:discrete}). We opt for this sparse formulation for several reasons: 1) empirical results demonstrate satisfactory performance even without these weights; 2) it simplifies the development of an efficient training algorithm; 3) excluding noisy data enhances the robustness of the model.

Given the discrete nature of the mask $m$, directly solving the bilevel optimization problem (\ref{eq:discrete}) is intractable due to its NP-hard nature. 
Hence, we adopt a continualization approach~\cite{DBLP:conf/icml/ZhouPZLCZ22} via probabilistic reparameterization to render gradient-based optimization feasible. We treat each mask $m_i$ as an independent binary random variable and transform the problem into the continuous probability space.
Specifically, we reparameterize $m_i$ as a Bernoulli random variable with probability $s_i$ for being $1$ and $1-s_i$ for being 0, i.e., $m_i \sim \text{Bern}(s_i)$, where $s_i \in [0, 1]$. Assuming independence among the variables $m_i$, the distribution function of $m$ becomes $p(m|s) = \prod ^n_{i=1}(s_i)^{m_i} (1-s_i)^{(1-m_i)}$. Thus, we control the selected size through the sum of probabilities $s_i$, since $\mathbb{E}_{m\sim p(m|s)}[\|m\|_0] =\sum_{i=1}^ns_i$. Consequently,
$\mathcal{C}$ can be relaxed into $\tilde{\mathcal{C}} = \{s_i : 0 \leq s_i \leq1, \|s\|_1 \leq K\}$. Finally, problem (\ref{eq:discrete}) naturally relaxes into the following:  

\begin{eqnarray} \label{eq:bern}
\underset{s\in\tilde{\mathcal{C}}}{\text{min}}\;\Psi(s)=\mathbb{E}_{p(m|s)}[\mathcal{R}(\mathcal{D},\theta^*(m))],\\
\text{s.t. } \theta^*(m) \in \underset{\theta}{\text{arg min}}\;\mathcal{L}(\mathcal{D},\theta; m) \notag
\end{eqnarray}

where $\tilde{\mathcal{C}} =\{s_i : 0 \leq s_i \leq 1,\|s\|_1 \leq K\}$ is the domain.

Several beneficial aspects of our formulation (\ref{eq:bern}) include:

\begin{enumerate}
\item Our formulation serves as a close relaxation (though not equivalent) of Problem (\ref{eq:discrete}). This is evident for the following reasons:

\begin{enumerate}
\item It is apparent that $\text{min}_{s\in\tilde{\mathcal{C}}}\Psi(s) \leq
\text{min}_{m\in\mathcal{C}}\Psi(m)$ since any deterministic binary mask $m$ can be represented as a specific stochastic one by setting $s_i$ to either 0 or 1.
\item Our constraint $\tilde{\mathcal{C}}$ induces sparsity on $s$ through the $l_1$-norm and the range $[0, 1]$, resulting in most components of the optimal $s$ being either 0 or 1. Therefore, our eventually learned stochastic weight is nearly deterministic.

\end{enumerate}
\item Due to the sparsity constraint, the size of the selected set in the inner loop, remains small, which greatly enhances the efficiency of optimizing $\theta^*$ (refer to details in Appendix).
\item As indicated in Eq. (\ref{eq:prob}), our outer objective $\Psi(s)$ is differentiable, enabling the utilization of general gradient-based methods for optimization.

\end{enumerate}

\subsection{Optimization Method}

Current bilevel optimization algorithms \cite{DBLP:conf/icml/Pedregosa16,DBLP:conf/icml/GrazziFPS20} typically incur high computational costs owing to the resource-intensive implicit differentiation inherent in their chain-rule-based gradient estimator. Specifically, if employed in our context, they commonly approximate the gradient in the following manner:

\begin{equation}
\nabla_s\Psi(s)\approx \nabla_s\theta^*(m)\nabla_{\theta}\mathcal{R}(\mathcal{D},\theta^*(m))
\end{equation}

Hence, they need to compute the implicit differentiation of the inner loop optimum, i.e, $\nabla_s\theta^*(m)$, which is expensive since they have to compute the inverse of a huge hessian matrix or unroll the backward propagation for multiple steps. Even though some efficient bilevel optimization algorithms have been proposed to alleviate the computational burden (for instance, \cite{DBLP:conf/aistats/LorraineVD20} adopted Neumann series to approximate the hessian inverse), the approximation is nevertheless time-consuming. 



The probabilistic formulation (\ref{eq:bern}) of the bilevel problem allows us to circumvent costly computations by computing the gradient using forward propagation instead of backward propagation. This can be illustrated by the following equations:
\begin{align} 
\nabla _s\Psi(s)
&=\nabla_s\mathbb{E}_{p(m|s)}[\mathcal{R}(\mathcal{D},\theta^*(m))] \nonumber\\
&= \nabla_s\int\mathcal{R}(\mathcal{D},\theta^*(m))p(m|s)dm \nonumber\\
&=\int\mathcal{R}(\mathcal{D},\theta^*(m))\frac{\nabla_sp(m|s)}{p(m|s)}p(m|s)dm \nonumber\\
&=\int\mathcal{R}(\mathcal{D},\theta^*(m))\nabla_s\text{ln}p(m|s)p(m|s)dm \nonumber\\
&=\mathbb{E}_{p(m|s)}[\mathcal{R}(\mathcal{D},\theta^*(m))\nabla_s\text{ln}p(m|s)] \label{eq:prob} 
\end{align}


This indicates that $\mathcal{R}(\mathcal{D},\theta^*(m)) \nabla_s\text {ln} p(m|s)$ serves as an unbiased stochastic gradient of $\nabla_s\Psi(s)$. Consequently, with the inner loop optimum $\theta^*(m)$ at hand, we can update $s$ (probability) via projected stochastic gradient descent:

\begin{equation}
s \leftarrow \mathcal{P}_{\tilde{\mathcal{C}}} (s- \eta\mathcal{R}(\mathcal{D},\theta^*(m)) \nabla_s \text{ln} p(m|s))
\end{equation}



It's evident that this approach does not entail any implicit differentiation, and its component $\mathcal{R}(\mathcal{D},\theta^*(m))$ can be computed through forward propagation. Additionally, $\text {ln} p(m|s)$ exhibits a straightforward form, and the projection possesses a closed-form solution \cite{DBLP:conf/icml/ZhouPZLCZ22} given the simplicity of the constraint $\tilde{\mathcal{C}}$. Consequently, we can efficiently update $s$.

Thus, we can tackle our bilevel optimization problem (\ref{eq:bern}) by alternately: 1) sampling $m$, i.e., a selected set, from $p(m|s)$ for the inner loop and training the model on this selected set to obtain $\theta^*(m)$; 2) updating the probability $s$. The details are shown in Algorithm \ref{al:1}.

\begin{table*}[!htb] 
\caption{ Accuracy and $\Delta\text{Suf}$ in Toxic comments (left) and CelebA datasets (right)} \label{table:complex}

    \begin{minipage}{.515\linewidth}

\scalebox{1.127}{
    \begin{tabular}{lll}
    
        \toprule
        \multicolumn{1}{c}{Toxic comments} & \multicolumn{1}{c}{Accuracy($\uparrow$)} &\makebox[2cm][c]{$\Delta\text{Suf}$$(\downarrow$)}\\

        \midrule
        \makebox[2cm][c]{ERM(I)}& 0.768$\pm$0.004 &  \makebox[2cm][c]{0.173$\pm$0.008} \\
        \midrule
        \makebox[2cm][c]{NUF(II)}&0.762$\pm$0.007& \makebox[2cm][c]{0.190$\pm$0.008} \\
        \midrule
        \makebox[2cm][c]{IPA(III)}& 0.745$\pm$0.007 &  \makebox[2cm][c]{0.091$\pm$0.012} \\
        \midrule
        \makebox[2cm][c]{AR(IV)}& 0.756$\pm$0.006 &  \makebox[2cm][c]{0.128$\pm$0.097} \\
        \midrule
        \makebox[2cm][c]{Ours(V)}& \bf0.763$\pm$0.004 &  \makebox[2cm][c]{\bf0.028$\pm$0.004} \\
        \midrule
        \makebox[2cm][c]{IRMv1(VI)}& 0.753$\pm$0.004 &  \makebox[2cm][c]{0.068$\pm$0.008} \\
        \bottomrule
    \end{tabular}
    }
    \end{minipage}
\begin{minipage}{.515\linewidth}    
\scalebox{1.127}{
   \begin{tabular}{lll}
    
        \toprule
        \multicolumn{1}{c}{CelebA} & \multicolumn{1}{c}{Accuracy($\uparrow$)} &\makebox[2cm][c]{$\Delta\text{Suf}$$(\downarrow$)}\\

        \midrule
        \makebox[2cm][c]{ERM(I)}& 0.956$\pm$0.005 &  \makebox[2cm][c]{0.210 $\pm$0.094} \\
        \midrule
        \makebox[2cm][c]{NUF(II)}&0.947$\pm$0.007 & \makebox[2cm][c]{0.104$\pm$0.004} \\
        \midrule
        \makebox[2cm][c]{IPA(III)}& 0.938$\pm$0.103 &  \makebox[2cm][c]{0.092$\pm$0.161} \\
        \midrule
        \makebox[2cm][c]{AR(IV)}& 0.950$\pm$0.012 &  \makebox[2cm][c]{0.197$\pm$0.007} \\
        \midrule
        \makebox[2cm][c]{Ours(V)}& \bf0.953$\pm$0.094 &  \makebox[2cm][c]{\bf0.045$\pm$0.004} \\
        \midrule
        \makebox[2cm][c]{IRMv1(VI)}& 0.946$\pm$0.009 &  \makebox[2cm][c]{0.088$\pm$0.007} \\
        \bottomrule
    \end{tabular}
 }
        \end{minipage} 
\end{table*}

\begin{table*}[!tb]
\vspace*{0.5 cm}
\caption{Accuracy and $\Delta\text{Suf}$ in Adult (left) and COMPAS datasets (right)} \label{table:tabular}

    \begin{minipage}{.515\linewidth}

\scalebox{1.127}{
    \begin{tabular}{lll}
    
        \toprule
        \multicolumn{1}{c}{Adult} & \multicolumn{1}{c}{Accuracy($\uparrow$)} &\makebox[2cm][c]{$\Delta\text{Suf}$$(\downarrow$)}\\

        \midrule
        \makebox[2cm][c]{ERM(I)}& 0.831$\pm$0.014 &  \makebox[2cm][c]{0.160$\pm$0.007} \\
        \midrule
        \makebox[2cm][c]{NUF(II)}&0.815$\pm$0.017 & \makebox[2cm][c]{0.068$\pm$0.015} \\
        \midrule
        \makebox[2cm][c]{IPA(III)}& 0.810$\pm$0.004 &  \makebox[2cm][c]{0.058$\pm$0.024} \\
        \midrule
        \makebox[2cm][c]{AR(IV)}& 0.820$\pm$0.023 &  \makebox[2cm][c]{0.230$\pm$0.014} \\
        \midrule
        \makebox[2cm][c]{Ours(V)}& \bf0.827$\pm$0.016 &  \makebox[2cm][c]{0.036$\pm$0.007} \\
        \midrule
        \makebox[2cm][c]{IRMv1(VI)}& 0.825$\pm$0.018 &  \makebox[2cm][c]{\bf0.032$\pm$0.012} \\
        \bottomrule
    \end{tabular}
    }
  
    \end{minipage}
\begin{minipage}{.515\linewidth}    
\scalebox{1.127}{
   \begin{tabular}{lll}
    
        \toprule
        \multicolumn{1}{c}{COMPAS} & \multicolumn{1}{c}{Accuracy($\uparrow$)} &\makebox[2cm][c]{$\Delta\text{Suf}$$(\downarrow$)}\\

        \midrule
        \makebox[2cm][c]{ERM(I)}& 0.652$\pm$0.024 &  \makebox[2cm][c]{0.276$\pm$0.094} \\
        \midrule
        \makebox[2cm][c]{NUF(II)}&0.633$\pm$0.032 & \makebox[2cm][c]{0.156$\pm$0.008} \\
        \midrule
        \makebox[2cm][c]{IPA(III)}& 0.647$\pm$0.017 &  \makebox[2cm][c]{0.097$\pm$0.009} \\
        \midrule
        \makebox[2cm][c]{AR(IV)}& \bf0.659$\pm$0.019 &  \makebox[2cm][c]{0.285$\pm$0.018} \\
        \midrule
        \makebox[2cm][c]{Ours(V)}& 0.647$\pm$0.004 &  \makebox[2cm][c]{\bf0.068$\pm$0.015} \\
        \midrule
        \makebox[2cm][c]{IRMv1(VI)}& 0.645$\pm$0.008 &  \makebox[2cm][c]{0.078$\pm$0.017} \\
        \bottomrule
    \end{tabular}
 }

        \end{minipage}
\vspace*{0.3cm}
        
\end{table*}


\begin{figure*}[ht]
\centering     
\subfloat[Accuracy vs. Noise ratio on Toxic comments]{\label{fig:noise1}\includegraphics[width=41.5mm,height=1.5in]{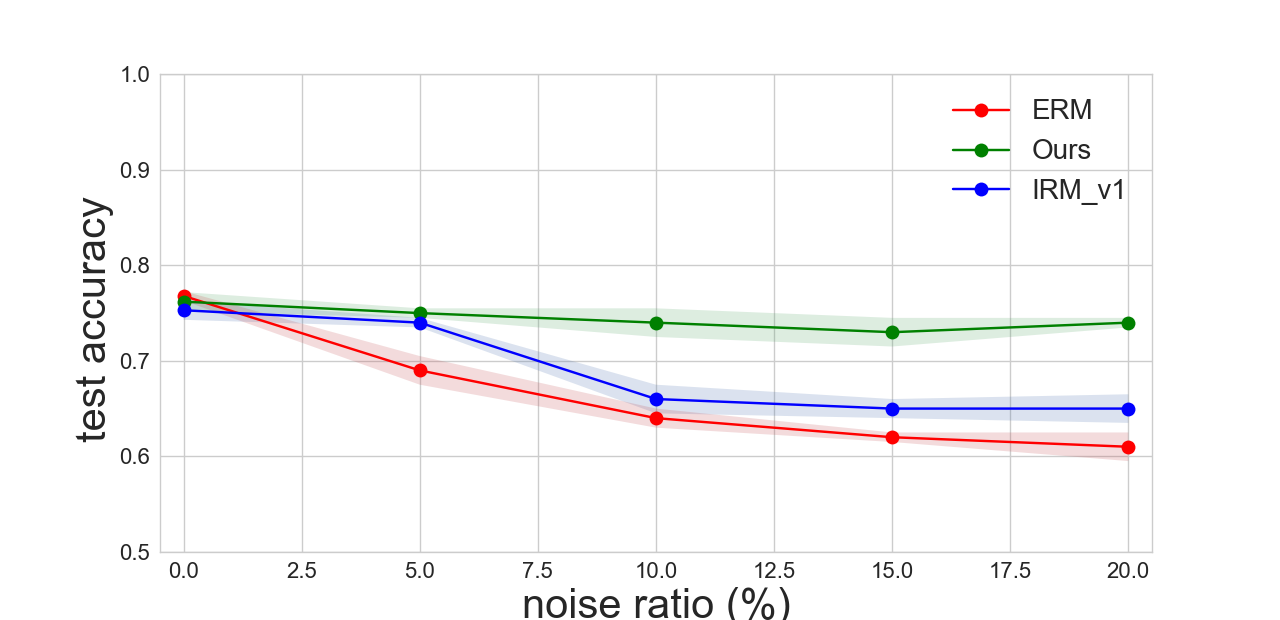}}
\hspace{3mm}%
\subfloat[Sufficiency gap vs. Noise ratio on toxic comments]{\label{fig:noise2}\includegraphics[width=41.5mm,height=1.5in]{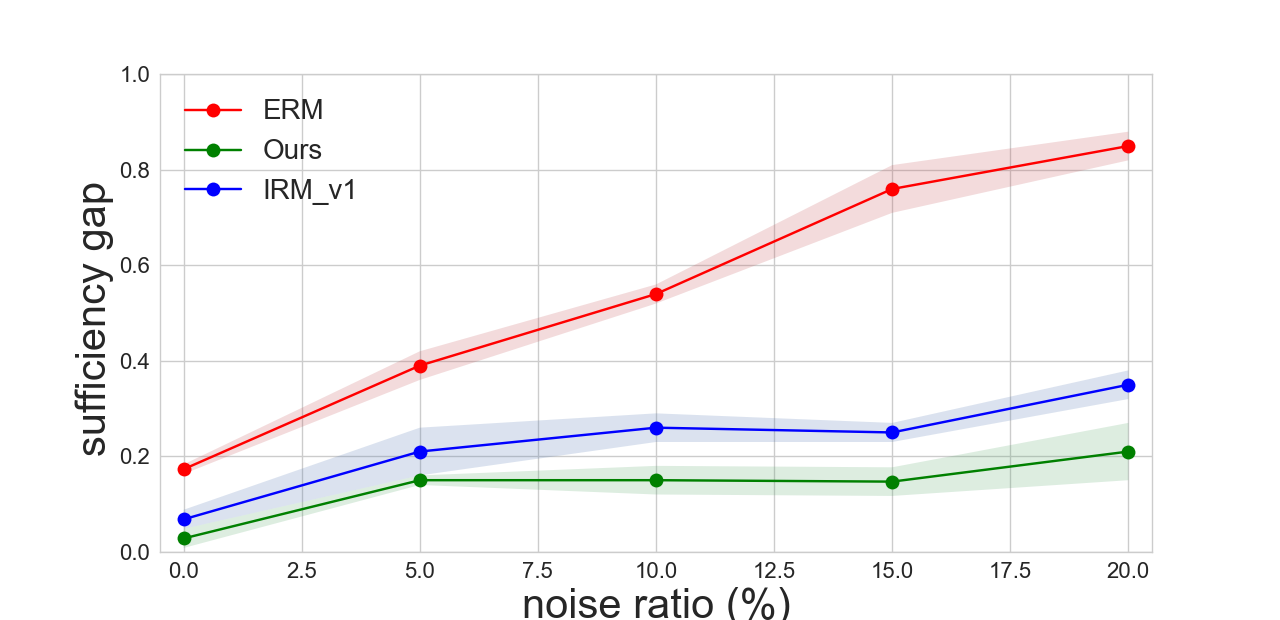}}
\hspace{3mm}%
\subfloat[Accuracy vs. Noise ratio on CelabA]{\label{fig:noise4}\includegraphics[width=41.5mm,height=1.5in]{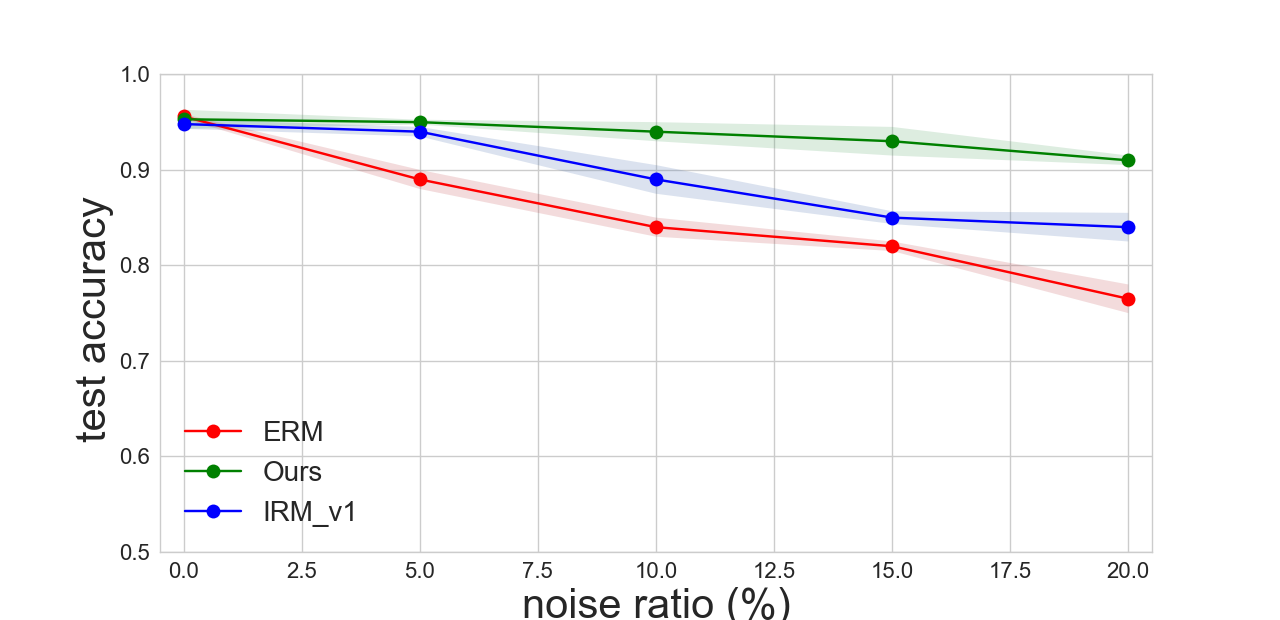}}
\hspace{3mm}%
\subfloat[Sufficiency gap vs. Noise ratio on CelabA]{\label{fig:noise5}\includegraphics[width=41.5mm,height=1.5in]{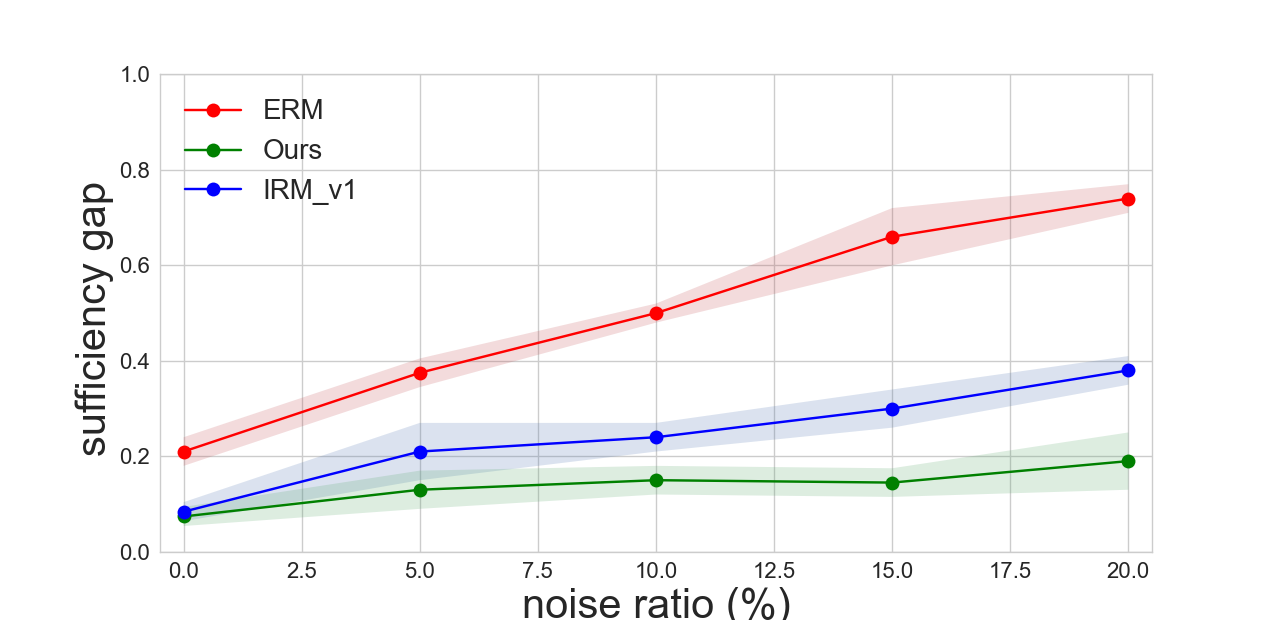}}
\vspace*{0.5cm}
\caption{Change of accuracy and sufficiency gap under different noise ratios on Toxic comments and CelebA datasets, which shows that our method is robust when the data label is noisy.
}
\hspace*{1cm}
 \label{fig:noise}
 \hspace*{2cm}
\end{figure*}

\hspace*{1cm}

\begin{algorithm}  

\caption{Reweighting for the Sufficiency Rule}
 \begin{algorithmic}[1] 
   
  \Require a neural network parameterized by $\theta$, a dataset $\mathcal{D}$, and a selected set size $K$.

  \State 
  Initiate probabilities $s^1$ as $\frac{K}{|\mathcal{D}|}\mathbf{1}$.
    
   \For {iteration $t$ of training, where $t$ is from 1 to $T$.}
   \State Sample $m$ based on the probability vector $s^t$.
   \State Continue training the inner loop until convergence achieved:
\par
   $\theta^*(m) \leftarrow \underset{\theta}{\text{argmin}} \hat{\mathcal{L}}(\theta;m)$
   \State Sample a mini-batch $\mathcal{K}$ from the dataset $\mathcal{D}$ :\par
   $\mathcal{K}=\{(x_1,y_1),...,(x_\mathcal{K},y_\mathcal{K})\}$ 
   \State Update $s$ according to $\theta^*(m)$ and $\mathcal{K}$. \par
   $s^{t+1} \leftarrow \mathcal{P}_{\tilde{\mathcal{C}}}(s^t-\eta\mathcal{R}_\mathcal{K}(\mathcal{D},\theta^*(m))\nabla_s\text{ln}p(m|s^t))$

   \EndFor
   
\State{Output: The selected set $\{(x_i, y_i): m_i= 1 \: \text{and} \: (x_i, y_i)\in\mathcal{D}\}$ where $m$ is sampled from $p(m|s^{T+1})$.}

   \end{algorithmic} \label{al:1}

   \end{algorithm}

\section{Experiments}\label{sec:exe}

We adopt the aforementioned sufficiency gap as the fair metric and accuracy as the metric for utility. Our neural network models are trained on an Intel(r) Core(TM) i7-8700 CPU. The networks in our experiments are built using the Pytorch package \cite{DBLP:conf/nips/PaszkeGMLBCKLGA19}.

\subsection{Baselines}

We compare our method with (I) Empirical Risk Minimization (\textbf{ERM}) which trains the model without considering fairness;
(II) No Utility-Cost Fairness via Data Reweighing (\textbf{NUF}) \cite{DBLP:conf/icml/LiL22a}; (III) Fair Representation Learning through Implicit Path Alignment (\textbf{IPA}) \cite{DBLP:conf/icml/ShuiC0W022}, an approach in the fair representation learning to achieve also the sufficiency rule; (IV) Adversarial Reweighting Guided by Wasserstein Distance for Bias Mitigation (\textbf{AR}) \cite{zhao2023adversarial}. Notably, the baseline (IV) is grounded in Demographic Parity (DP), illustrating their general incompatibility with addressing the sufficiency rule. Additionally, we include the original Invariant Risk Minimization regularization \cite{DBLP:journals/corr/abs-1907-02893}, denoted as IRMv1, which incorporates a gradient penalty to encourage invariance across different groups. Even though it is designed for another purpose, as shown earlier in Section \ref{sec:method}, it has potential to address fairness to reach the sufficiency. Results are averaged over five repetitions. Further experimental results are provided in Appendix.

\subsection{Datasets and Experiment Setups}\label{sec:data}


The \textbf{toxic comments dataset} \cite{WB:2014} presents a binary classification challenge in natural language processing (NLP), aiming to determine whether a comment exhibits toxicity. Originally, the labeling process for this dataset is not binary due to involvement from multiple annotators, leading to potential discrepancies. To address this, we adopt a straightforward strategy where a comment is classified as toxic if at least one annotator marks it as such, similar to the approach in \cite{DBLP:conf/icml/ShuiC0W022}. Notably, some comments in this dataset are annotated with identity attributes such as gender and race. It has been observed that the race attribute correlates with the toxicity label, posing a risk of predictive discrimination. Therefore, we designate race as the protected feature and specifically focus on two sub-groups: Black and Asian. For computational efficiency, we begin by leveraging a pre-trained BERT model \cite{DBLP:conf/naacl/DevlinCLT19} to extract word embeddings, resulting in vectors of 748 dimensions.


The \textbf{CelebA dataset} \cite{liu2015faceattributes} comprises approximately 200K images featuring celebrity faces, each associated with 40 human-annotated binary attributes such as gender, hair color, and age. For our experiment, we randomly partitioned the dataset, selecting approximately 82K images for training and 18K for validation. We employed the ResNet-18 architecture \cite{DBLP:conf/cvpr/HeZRS16}, pre-trained on ImageNet \cite{DBLP:conf/cvpr/DengDSLL009}, omitting the final fully-connected layer to obtain embeddings of 512 dimensions for simplicity. Within the CelebA dataset, our specific task involves predicting hair color (\{blond, dark\}) based on the image input. Notably, the gender attribute (\{male, female\}) is correlated with hair color.

For experiments on tabular data, we use the \textbf{Adult dataset} \cite{kohavi1996} and the \textbf{COMPAS dataset} \cite{mattu} (For more details of the datasets, please refer to Appendix). Adult dataset used personal information such as
education level and working hours per week to predict whether an individual earns more or less than \$50,000 per year. We use gender as the sensitive feature in Adult dataset. COMPAS dataset is a popular
commercial algorithm used by judges and parole officers for scoring criminal defendant’s likelihood of reoffending. We use race as the sensitive feature. Here we report the sufficiency gap between two sub-groups of African American and Caucasian even though that ethnicity group is a multi-categorical feature. Please refer to Appendix for more data and training details.

\subsection{Analysis}
\subsubsection{Performance Comparison}

In Table \ref{table:complex} and Table \ref{table:tabular}, we present the accuracy and sufficiency gap metrics. Notably, we observe that the Demographic Parity (DP) based fair method (IV) is incompatible with the sufficiency rule, as evidenced by its tendency to increase $\Delta\text{Suf}$ even surpassing that of ERM. On the other hand, baselines (III, VI), which aim to track the sufficiency rule, exhibit improved sufficiency gap $\Delta\text{Suf}$ with comparable accuracy, albeit inferior to our approach in Table \ref{table:complex}. This discrepancy may stem from an overparameterization issue, as previously discussed. Our method consistently demonstrates a superior Accuracy-Fairness trade-off, significantly enhancing sufficiency without substantial accuracy loss. We observe a similar performance pattern on tabular datasets (Table \ref{table:tabular}). However, the performance drop of baselines (III, VI) is less pronounced. This discrepancy may be attributed to the comparatively smaller DNN models utilized in training on tabular datasets, which are less susceptible to overparameterization compared to the toxic comments dataset and CelebA dataset.

\begin{figure*}[!t]
\captionsetup[subfigure]{labelformat=empty}
\centering     
\subfloat[]
{\includegraphics[width=41.5mm,height=1.5in]{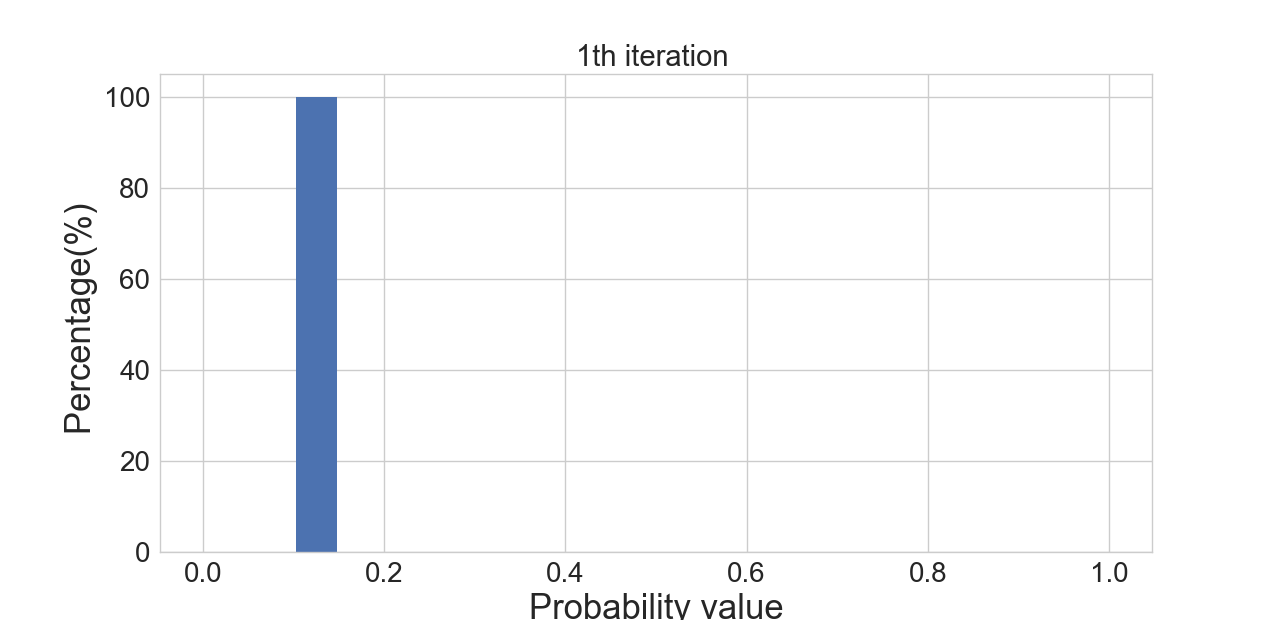}}
\hspace{3mm}%
\subfloat[]{\includegraphics[width=41.5mm,height=1.5in]{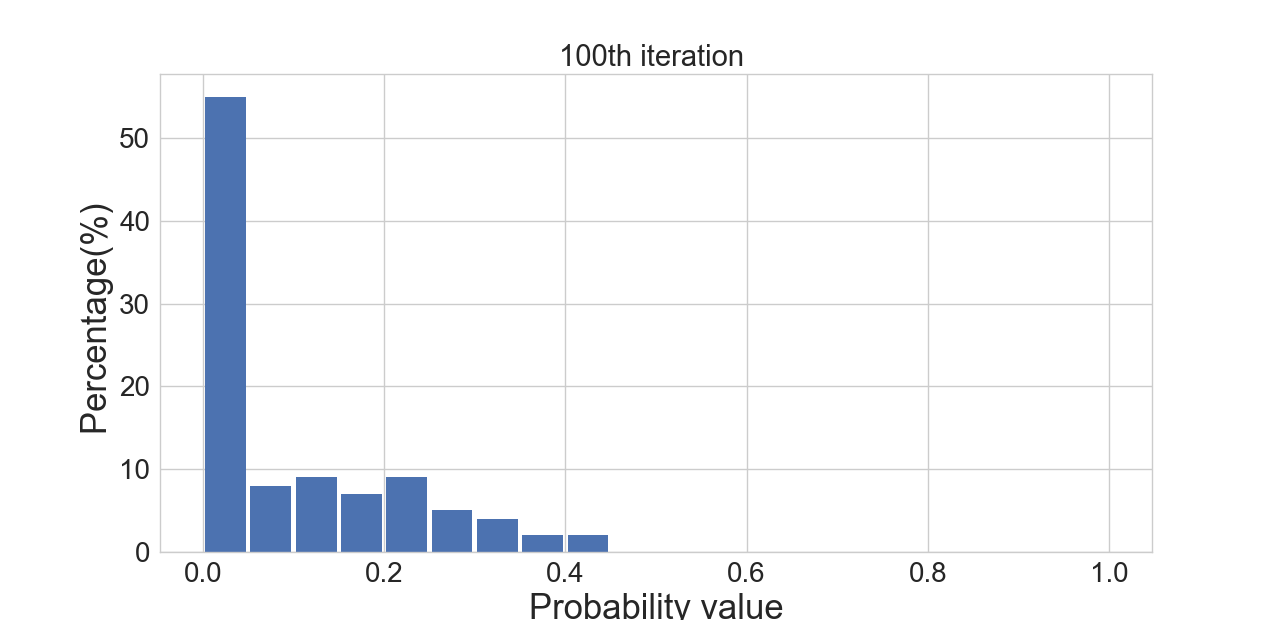}}
\hspace{3mm}%
\subfloat[]{\includegraphics[width=41.5mm,height=1.5in]{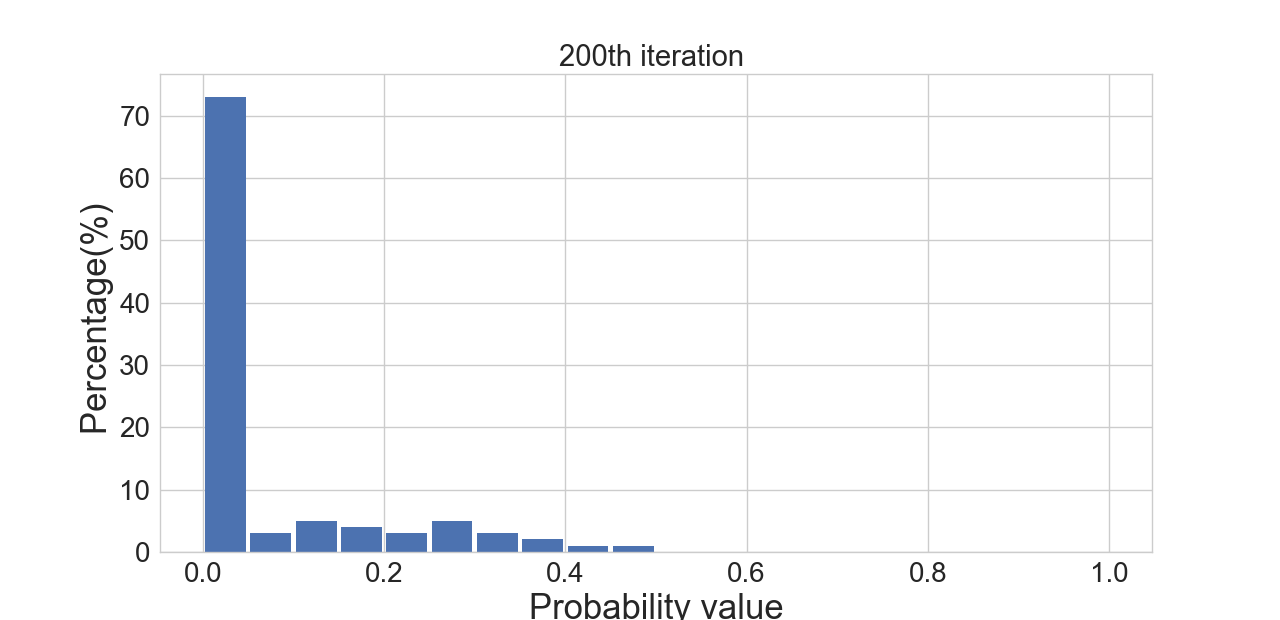}}
\hspace{3mm}%
\subfloat[]{\includegraphics[width=41.5mm,height=1.5in]{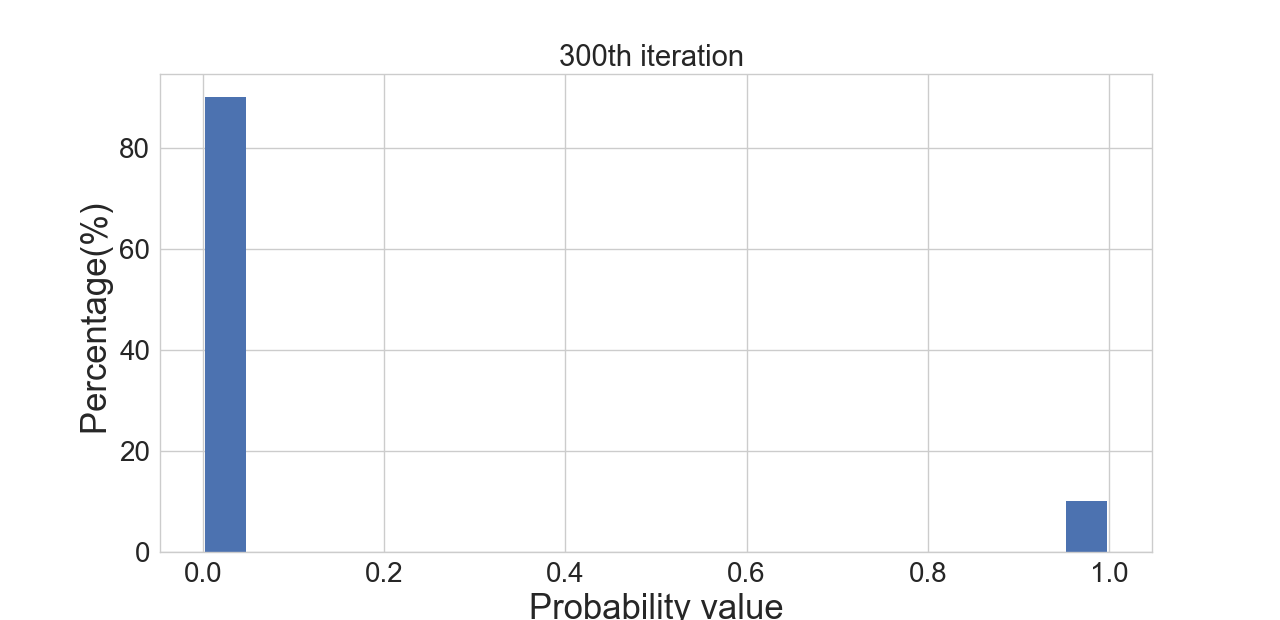}}
\caption{The evolution of probability score distribution during the search process reveals a trend where most probabilities tend to converge towards either 0 or 1. This convergence ultimately leads to deterministic weights and convergence of the algorithm.}
 \label{fig:s}
  \vspace*{0.7cm}
\end{figure*}

\subsubsection{Robustness with Noisy Data}

We extend our experimentation to scenarios where the dataset incorporates corrupted labels, aiming to demonstrate the robustness of our approach. Following the model configuration outlined in Section \ref{sec:data}, we introduce symmetric noise \cite{9729424} into the dataset. Notably, as illustrated in Figure \ref{fig:noise}, our method exhibits robustness towards variations in the dataset's label quality, as evidenced by consistent performance in both accuracy and sufficiency gap metrics. This robustness can be attributed to the comprehensive information assimilated through iterative sampling, leading to the construction of the final weight vector $w$. Essentially, the sparsity induced by our method facilitates the elimination of noisy data samples, thus preserving the model's effectiveness.

\subsubsection{Sensitivity to Choices of K}

\begin{figure} [!htbp]
\centering     
\subfloat[size $K$ for Toxic comments dataset]{\label{fig:noise7}\includegraphics[width=65.5mm,height=1.8in]{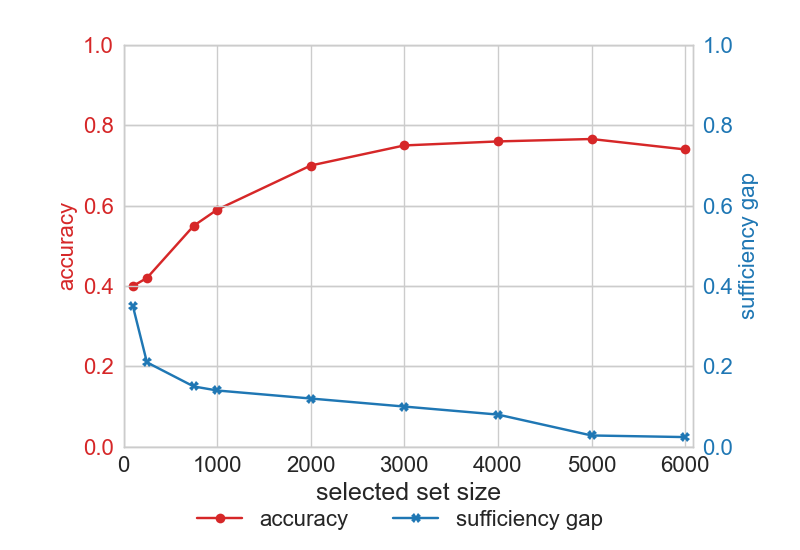}}
\hspace{\floatsep}
\subfloat[size $K$ for CelebA dataset]{\label{fig:noise8}\includegraphics[width=65.5mm,height=1.8in]{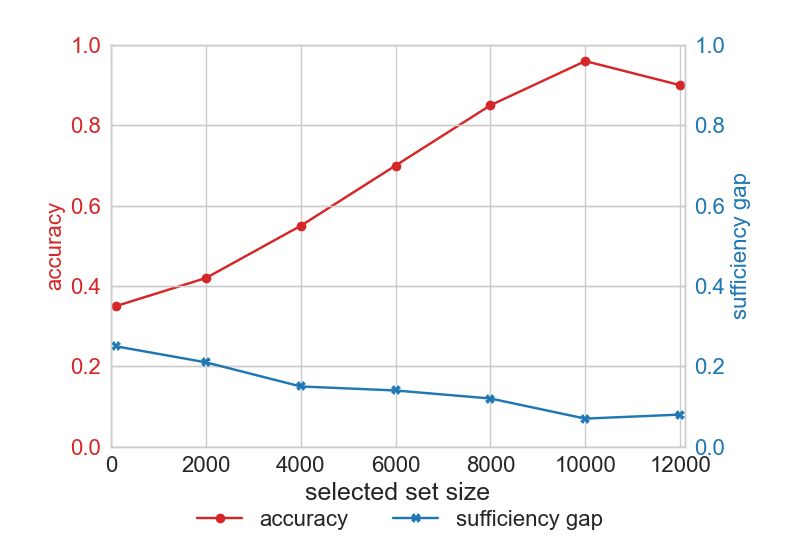}}
\vspace*{0.3cm}
\caption{Choices of $K$ (selected set size). The size is set to 5000 for Toxic comments and 10000 for
CelebA. }
 \label{fig:size}
 \vspace*{0.7cm}
\end{figure}

The selected sizes for the Toxic comments and CelebA experiments are 5000 and 10000, respectively, as shown in Figure \ref{fig:size}. As the selected sizes increase, we observe an improvement in both accuracy and sufficiency gap performance. Yet, beyond a certain threshold, this improvement plateaus, aligning with the corset concept \cite{DBLP:conf/icml/MirzasoleimanBL20}. Corset theory suggests that there exists a small subset capable of summarizing the larger dataset effectively. Training exclusively on this condensed set yields competitive performance compared to training on the entire dataset.

\subsubsection{Convergence of Probabilities during Search} 

A simplified approach is taken by selecting 1000 samples from a larger pool of 10000 training data instances (CelebA). Figure \ref{fig:s} illustrates the evolution of probability distributions throughout the search process. Initially, all sample probabilities are uniformly distributed at 0.1. Over the course of the search, most of these probabilities tend to converge towards either 0 or 1, indicative of diminishing uncertainty. Consequently, a sparse mask with minimal variance is formed, reflecting a nearly deterministic pattern in weight assignment. This trend ultimately leads to the establishment of deterministic weights, signifying algorithmic convergence.

\subsubsection{Gradual Change of Group Weights
} 

\begin{figure}  [ht]
    \centering
    \includegraphics[width=.5\textwidth,height=2.4in]{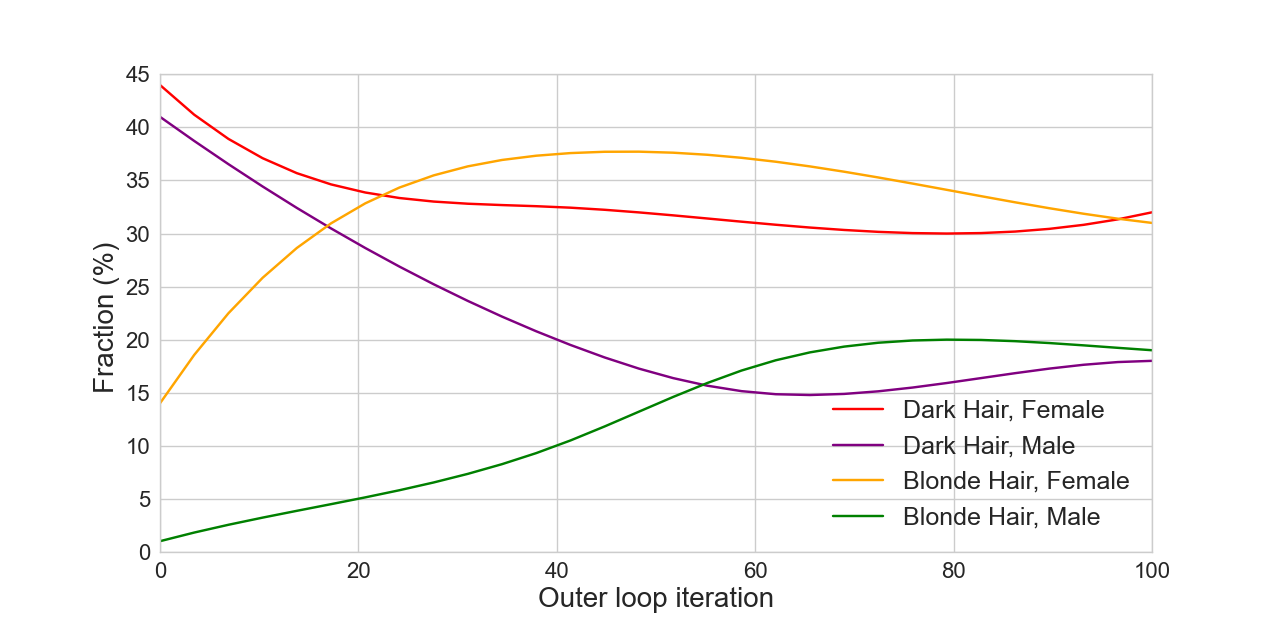}
    \vspace*{0.5 cm}
    \caption{The fluctuation in the distribution of group weight fractions for ResNet-18 on the CelebA dataset is notable. Specifically, there's a shift to 20\% for both the (Male, Blond Hair) and (Male, Dark Hair) gourps. Similarly, the (Female, Dark Hair) and (Female, Blond Hair) groups see their fractions adjusted to 30\%. These observations suggest that our methodology is capable of autonomously adapting weight fractions across various (sub)-groups.}
    \label{fig:imbalance}
    \vspace*{0.7cm}
\end{figure}

In Figure \ref{fig:imbalance}, we depict the training dynamics of sample weight fractions from our CelebA experiment. Initially, all sample weights are uniformly set to 1. The weight fraction of the (Male, Blond Hair) group begins at a mere 0.085\%. Following 100 iterations of updates, this fraction gradually increases to approximately 20\%. Simultaneously, the weight fraction of the (Male, Dark Hair) group decreases to approximately 20\%, while both the (Female, Dark Hair) and (Female, Blond Hair) groups stabilize at approximately 30\%. Interestingly, in \cite{chai2022}, the assumption is that bias is introduced due to under-representation of the minority groups, hence, they upweight/downweight sensitive groups to the same importance level. Figure \ref{fig:imbalance} demonstrates that even though we do not constrain on the group level importance, somehow, our method possesses the ability to dynamically adjust the weight fraction of (sub)-groups automatically.

\section{Discussion and Conclusion}
We presented a model agnostic sample reweighing method to achieve the sufficiency rule of fairness. We formulated this problem as a bilevel optimization to learn sample weights. We further enhance our method with sparsity constraints to improve training speed. Then, we analyzed the sufficiency gap and prediction accuracy of the reweighting algorithm, demonstrating its superior performance over state-of-the-art approaches. The empirical results also show that our method is robust towards noisy labels. One limitation of our framework is that the overall performance of IRMv1, in terms of test accuracy, consistently improves when there is a significant difference between the training environments \cite{DBLP:journals/corr/abs-2004-05007}. This suggests that, in terms of fairness, IRMv1 is more effective when there is greater disparity between the sensitive sub-groups.

\ack This work has received funding from the European Union’s Horizon 2020 research and innovation programme under Marie Sklodowska-Curie Actions (grant agreement number 860630) for the project “NoBIAS - Artificial Intelligence without Bias”.




\bibliography{mybibfile}

\begin{thebibliography}{46}
\providecommand{\natexlab}[1]{#1}
\providecommand{\url}[1]{\texttt{#1}}
\expandafter\ifx\csname urlstyle\endcsname\relax
  \providecommand{\doi}[1]{doi: #1}\else
  \providecommand{\doi}{doi: \begingroup \urlstyle{rm}\Url}\fi

\bibitem[Ahmad et~al.(2020)Ahmad, Patel, Eckert, Kumar, and Teredesai]{DBLP:conf/kdd/AhmadPEKT20}
M.~A. Ahmad, A.~Patel, C.~Eckert, V.~Kumar, and A.~Teredesai.
\newblock Fairness in machine learning for healthcare.
\newblock In \emph{{KDD}}, pages 3529--3530. {ACM}, 2020.

\bibitem[Arjovsky et~al.(2019)Arjovsky, Bottou, Gulrajani, and Lopez{-}Paz]{DBLP:journals/corr/abs-1907-02893}
M.~Arjovsky, L.~Bottou, I.~Gulrajani, and D.~Lopez{-}Paz.
\newblock Invariant risk minimization.
\newblock \emph{CoRR}, abs/1907.02893, 2019.

\bibitem[Barocas et~al.(2023)Barocas, Hardt, and Narayanan]{barocas-hardt-narayanan}
S.~Barocas, M.~Hardt, and A.~Narayanan.
\newblock \emph{Fairness and Machine Learning: Limitations and Opportunities}.
\newblock MIT Press, 2023.

\bibitem[Bühlmann(2018)]{bühlmann2018invariance}
P.~Bühlmann.
\newblock Invariance, causality and robustness, 2018.

\bibitem[Castelnovo et~al.(2022)Castelnovo, Crupi, Greco, Regoli, Penco, and Cosentini]{Castelnovo_2022}
A.~Castelnovo, R.~Crupi, G.~Greco, D.~Regoli, I.~G. Penco, and A.~C. Cosentini.
\newblock A clarification of the nuances in the fairness metrics landscape.
\newblock \emph{Scientific Reports}, 12, Mar. 2022.
\newblock ISSN 2045-2322.

\bibitem[Chai and Wang(2022)]{chai2022}
J.~Chai and X.~Wang.
\newblock Fairness with {{Adaptive Weights}}.
\newblock In \emph{Proceedings of the 39th {{International Conference}} on {{Machine Learning}}}, pages 2853--2866. {PMLR}, 2022.

\bibitem[Chang et~al.(2020)Chang, Zhang, Yu, and Jaakkola]{DBLP:conf/icml/ChangZYJ20}
S.~Chang, Y.~Zhang, M.~Yu, and T.~S. Jaakkola.
\newblock Invariant rationalization.
\newblock In \emph{{ICML}}, volume 119 of \emph{Proceedings of Machine Learning Research}, pages 1448--1458. {PMLR}, 2020.

\bibitem[Chen et~al.(2022)Chen, Rosenfeld, Sellke, Ma, and Risteski]{DBLP:conf/nips/ChenRS0R22}
Y.~Chen, E.~Rosenfeld, M.~Sellke, T.~Ma, and A.~Risteski.
\newblock Iterative feature matching: Toward provable domain generalization with logarithmic environments.
\newblock In \emph{NeurIPS}, 2022.

\bibitem[Cherepanova et~al.(2021)Cherepanova, Nanda, Goldblum, Dickerson, and Goldstein]{DBLP:journals/corr/abs-2102-06764}
V.~Cherepanova, V.~Nanda, M.~Goldblum, J.~P. Dickerson, and T.~Goldstein.
\newblock Technical challenges for training fair neural networks.
\newblock \emph{CoRR}, abs/2102.06764, 2021.

\bibitem[Choe et~al.(2020{\natexlab{a}})Choe, Ham, and Park]{DBLP:journals/corr/abs-2004-05007}
Y.~J. Choe, J.~Ham, and K.~Park.
\newblock An empirical study of invariant risk minimization.
\newblock \emph{CoRR}, abs/2004.05007, 2020{\natexlab{a}}.

\bibitem[Choe et~al.(2020{\natexlab{b}})Choe, Ham, and Park]{choe2020empirical}
Y.~J. Choe, J.~Ham, and K.~Park.
\newblock An empirical study of invariant risk minimization, 2020{\natexlab{b}}.

\bibitem[Chouldechova(2017)]{DBLP:journals/bigdata/Chouldechova17}
A.~Chouldechova.
\newblock Fair prediction with disparate impact: {A} study of bias in recidivism prediction instruments.
\newblock \emph{Big Data}, 5\penalty0 (2):\penalty0 153--163, 2017.

\bibitem[Chzhen et~al.(2020)Chzhen, Denis, Hebiri, Oneto, and Pontil]{DBLP:conf/nips/ChzhenDHOP20}
E.~Chzhen, C.~Denis, M.~Hebiri, L.~Oneto, and M.~Pontil.
\newblock Fair regression with wasserstein barycenters.
\newblock In \emph{NeurIPS}, 2020.

\bibitem[Deng et~al.(2009)Deng, Dong, Socher, Li, Li, and Fei{-}Fei]{DBLP:conf/cvpr/DengDSLL009}
J.~Deng, W.~Dong, R.~Socher, L.~Li, K.~Li, and L.~Fei{-}Fei.
\newblock Imagenet: {A} large-scale hierarchical image database.
\newblock In \emph{{CVPR}}, pages 248--255. {IEEE} Computer Society, 2009.

\bibitem[Devlin et~al.(2019)Devlin, Chang, Lee, and Toutanova]{DBLP:conf/naacl/DevlinCLT19}
J.~Devlin, M.~Chang, K.~Lee, and K.~Toutanova.
\newblock {BERT:} pre-training of deep bidirectional transformers for language understanding.
\newblock In \emph{{NAACL-HLT} {(1)}}, pages 4171--4186. Association for Computational Linguistics, 2019.

\bibitem[Grazzi et~al.(2020)Grazzi, Franceschi, Pontil, and Salzo]{DBLP:conf/icml/GrazziFPS20}
R.~Grazzi, L.~Franceschi, M.~Pontil, and S.~Salzo.
\newblock On the iteration complexity of hypergradient computation.
\newblock In \emph{{ICML}}, volume 119 of \emph{Proceedings of Machine Learning Research}, pages 3748--3758. {PMLR}, 2020.

\bibitem[Gulrajani and Lopez{-}Paz(2021)]{DBLP:conf/iclr/GulrajaniL21}
I.~Gulrajani and D.~Lopez{-}Paz.
\newblock In search of lost domain generalization.
\newblock In \emph{{ICLR}}. OpenReview.net, 2021.

\bibitem[Hardt et~al.(2016)Hardt, Price, and Srebro]{DBLP:conf/nips/HardtPNS16}
M.~Hardt, E.~Price, and N.~Srebro.
\newblock Equality of opportunity in supervised learning.
\newblock In \emph{{NIPS}}, pages 3315--3323, 2016.

\bibitem[He et~al.(2016)He, Zhang, Ren, and Sun]{DBLP:conf/cvpr/HeZRS16}
K.~He, X.~Zhang, S.~Ren, and J.~Sun.
\newblock Deep residual learning for image recognition.
\newblock In \emph{{CVPR}}, pages 770--778. {IEEE} Computer Society, 2016.

\bibitem[Jigsaw(2018)]{WB:2014}
Jigsaw.
\newblock Toxic comment classification challenge, 2018.
\newblock URL \url{https://www.kaggle.com/c/jigsaw-toxic-comment-classification-challenge}.

\bibitem[Kohavi(1996)]{kohavi1996}
R.~Kohavi.
\newblock Scaling up the accuracy of {{Naive-Bayes}} classifiers: A decision-tree hybrid.
\newblock In \emph{Proceedings of the {{Second International Conference}} on {{Knowledge Discovery}} and {{Data Mining}}}, {{KDD}}'96, pages 202--207, {Portland, Oregon}, Aug. 1996. {AAAI Press}.

\bibitem[Krasanakis et~al.(2018)Krasanakis, Xioufis, Papadopoulos, and Kompatsiaris]{DBLP:conf/www/KrasanakisXPK18}
E.~Krasanakis, E.~S. Xioufis, S.~Papadopoulos, and Y.~Kompatsiaris.
\newblock Adaptive sensitive reweighting to mitigate bias in fairness-aware classification.
\newblock In \emph{{WWW}}, pages 853--862. {ACM}, 2018.

\bibitem[Krueger et~al.(2021)Krueger, Caballero, Jacobsen, Zhang, Binas, Zhang, Priol, and Courville]{DBLP:conf/icml/KruegerCJ0BZPC21}
D.~Krueger, E.~Caballero, J.~Jacobsen, A.~Zhang, J.~Binas, D.~Zhang, R.~L. Priol, and A.~C. Courville.
\newblock Out-of-distribution generalization via risk extrapolation (rex).
\newblock In \emph{{ICML}}, volume 139 of \emph{Proceedings of Machine Learning Research}, pages 5815--5826. {PMLR}, 2021.

\bibitem[Li and Liu(2022)]{DBLP:conf/icml/LiL22a}
P.~Li and H.~Liu.
\newblock Achieving fairness at no utility cost via data reweighing with influence.
\newblock In \emph{{ICML}}, volume 162 of \emph{Proceedings of Machine Learning Research}, pages 12917--12930. {PMLR}, 2022.

\bibitem[Lin et~al.(2022)Lin, Dong, Wang, and Zhang]{DBLP:conf/cvpr/LinDWZ22}
Y.~Lin, H.~Dong, H.~Wang, and T.~Zhang.
\newblock Bayesian invariant risk minimization.
\newblock In \emph{{CVPR}}, pages 16000--16009. {IEEE}, 2022.

\bibitem[Liu et~al.(2019)Liu, Simchowitz, and Hardt]{DBLP:conf/icml/LiuSH19}
L.~T. Liu, M.~Simchowitz, and M.~Hardt.
\newblock The implicit fairness criterion of unconstrained learning.
\newblock In \emph{{ICML}}, volume~97 of \emph{Proceedings of Machine Learning Research}, pages 4051--4060. {PMLR}, 2019.

\bibitem[Liu et~al.(2015)Liu, Luo, Wang, and Tang]{liu2015faceattributes}
Z.~Liu, P.~Luo, X.~Wang, and X.~Tang.
\newblock Deep learning face attributes in the wild.
\newblock In \emph{Proceedings of International Conference on Computer Vision (ICCV)}, December 2015.

\bibitem[Lorraine et~al.(2020)Lorraine, Vicol, and Duvenaud]{DBLP:conf/aistats/LorraineVD20}
J.~Lorraine, P.~Vicol, and D.~Duvenaud.
\newblock Optimizing millions of hyperparameters by implicit differentiation.
\newblock In \emph{{AISTATS}}, volume 108 of \emph{Proceedings of Machine Learning Research}, pages 1540--1552. {PMLR}, 2020.

\bibitem[Madras et~al.(2018)Madras, Creager, Pitassi, and Zemel]{DBLP:conf/icml/MadrasCPZ18}
D.~Madras, E.~Creager, T.~Pitassi, and R.~S. Zemel.
\newblock Learning adversarially fair and transferable representations.
\newblock In \emph{{ICML}}, volume~80 of \emph{Proceedings of Machine Learning Research}, pages 3381--3390. {PMLR}, 2018.

\bibitem[Mattu et~al.(2016)Mattu, Angwin, Kirchner, Surya, and Larson]{mattu}
Mattu, J.~Angwin, L.~Kirchner, Surya, and J.~Larson.
\newblock How {{We Analyzed}} the {{COMPAS Recidivism Algorithm}}, 2016.

\bibitem[Mirzasoleiman et~al.(2020)Mirzasoleiman, Bilmes, and Leskovec]{DBLP:conf/icml/MirzasoleimanBL20}
B.~Mirzasoleiman, J.~A. Bilmes, and J.~Leskovec.
\newblock Coresets for data-efficient training of machine learning models.
\newblock In \emph{{ICML}}, volume 119 of \emph{Proceedings of Machine Learning Research}, pages 6950--6960. {PMLR}, 2020.

\bibitem[Obermeyer et~al.(2019)Obermeyer, Powers, Vogeli, and Mullainathan]{Obermeyer2019}
Z.~Obermeyer, B.~Powers, C.~Vogeli, and S.~Mullainathan.
\newblock Dissecting racial bias in an algorithm used to manage the health of populations.
\newblock \emph{Science}, 366:\penalty0 447--453, 2019.

\bibitem[Paszke et~al.(2019)Paszke, Gross, Massa, Lerer, Bradbury, Chanan, Killeen, Lin, Gimelshein, Antiga, Desmaison, K{\"{o}}pf, Yang, DeVito, Raison, Tejani, Chilamkurthy, Steiner, Fang, Bai, and Chintala]{DBLP:conf/nips/PaszkeGMLBCKLGA19}
A.~Paszke, S.~Gross, F.~Massa, A.~Lerer, J.~Bradbury, G.~Chanan, T.~Killeen, Z.~Lin, N.~Gimelshein, L.~Antiga, A.~Desmaison, A.~K{\"{o}}pf, E.~Z. Yang, Z.~DeVito, M.~Raison, A.~Tejani, S.~Chilamkurthy, B.~Steiner, L.~Fang, J.~Bai, and S.~Chintala.
\newblock Pytorch: An imperative style, high-performance deep learning library.
\newblock In \emph{NeurIPS}, pages 8024--8035, 2019.

\bibitem[Pedregosa(2016)]{DBLP:conf/icml/Pedregosa16}
F.~Pedregosa.
\newblock Hyperparameter optimization with approximate gradient.
\newblock In \emph{{ICML}}, volume~48 of \emph{{JMLR} Workshop and Conference Proceedings}, pages 737--746. JMLR.org, 2016.

\bibitem[Peters et~al.(2015)Peters, Bühlmann, and Meinshausen]{peters2015causal}
J.~Peters, P.~Bühlmann, and N.~Meinshausen.
\newblock Causal inference using invariant prediction: identification and confidence intervals, 2015.

\bibitem[Rosenfeld et~al.(2021)Rosenfeld, Ravikumar, and Risteski]{DBLP:conf/iclr/RosenfeldRR21}
E.~Rosenfeld, P.~K. Ravikumar, and A.~Risteski.
\newblock The risks of invariant risk minimization.
\newblock In \emph{{ICLR}}. OpenReview.net, 2021.

\bibitem[Shui et~al.(2022)Shui, Chen, Li, Wang, and Gagn{\'{e}}]{DBLP:conf/icml/ShuiC0W022}
C.~Shui, Q.~Chen, J.~Li, B.~Wang, and C.~Gagn{\'{e}}.
\newblock Fair representation learning through implicit path alignment.
\newblock In \emph{{ICML}}, volume 162 of \emph{Proceedings of Machine Learning Research}, pages 20156--20175. {PMLR}, 2022.

\bibitem[Song et~al.(2023)Song, Kim, Park, Shin, and Lee]{9729424}
H.~Song, M.~Kim, D.~Park, Y.~Shin, and J.-G. Lee.
\newblock Learning from noisy labels with deep neural networks: A survey.
\newblock \emph{IEEE Transactions on Neural Networks and Learning Systems}, 34\penalty0 (11):\penalty0 8135--8153, 2023.

\bibitem[Song et~al.(2019)Song, Kalluri, Grover, Zhao, and Ermon]{DBLP:conf/aistats/SongKGZE19}
J.~Song, P.~Kalluri, A.~Grover, S.~Zhao, and S.~Ermon.
\newblock Learning controllable fair representations.
\newblock In \emph{{AISTATS}}, volume~89 of \emph{Proceedings of Machine Learning Research}, pages 2164--2173. {PMLR}, 2019.

\bibitem[Xie et~al.(2020)Xie, Chen, Liu, and Li]{DBLP:journals/corr/abs-2006-07544}
C.~Xie, F.~Chen, Y.~Liu, and Z.~Li.
\newblock Risk variance penalization: From distributional robustness to causality.
\newblock \emph{CoRR}, abs/2006.07544, 2020.

\bibitem[Xu and Jaakkola(2021)]{DBLP:journals/corr/abs-2110-09940}
Y.~Xu and T.~S. Jaakkola.
\newblock Learning representations that support robust transfer of predictors.
\newblock \emph{CoRR}, abs/2110.09940, 2021.

\bibitem[Zemel et~al.(2013)Zemel, Wu, Swersky, Pitassi, and Dwork]{DBLP:conf/icml/ZemelWSPD13}
R.~S. Zemel, Y.~Wu, K.~Swersky, T.~Pitassi, and C.~Dwork.
\newblock Learning fair representations.
\newblock In \emph{{ICML} {(3)}}, volume~28 of \emph{{JMLR} Workshop and Conference Proceedings}, pages 325--333. JMLR.org, 2013.

\bibitem[Zhao et~al.(2023{\natexlab{a}})Zhao, Broelemann, Ruggieri, and Kasneci]{DBLP:conf/bigdataconf/ZhaoBRK23}
X.~Zhao, K.~Broelemann, S.~Ruggieri, and G.~Kasneci.
\newblock Causal fairness-guided dataset reweighting using neural networks.
\newblock In \emph{{IEEE} Big Data}, pages 1386--1394. {IEEE}, 2023{\natexlab{a}}.

\bibitem[Zhao et~al.(2023{\natexlab{b}})Zhao, Fabbrizzi, Lobo, Ghodsi, Broelemann, Staab, and Kasneci]{zhao2023adversarial}
X.~Zhao, S.~Fabbrizzi, P.~R. Lobo, S.~Ghodsi, K.~Broelemann, S.~Staab, and G.~Kasneci.
\newblock Adversarial reweighting guided by wasserstein distance for bias mitigation, 2023{\natexlab{b}}.

\bibitem[Zhao et~al.(2024)Zhao, Broelemann, Ruggieri, and Kasnecic]{zhao2024ecai}
X.~Zhao, K.~Broelemann, S.~Ruggieri, and G.~Kasnecic.
\newblock Enhancing fairness through reweighting: A path to attain the sufficiency rule.
\newblock \emph{arXiv preprint arXiv:1409.0473}, 2024.

\bibitem[Zhou et~al.(2022)Zhou, Pi, Zhang, Lin, Chen, and Zhang]{DBLP:conf/icml/ZhouPZLCZ22}
X.~Zhou, R.~Pi, W.~Zhang, Y.~Lin, Z.~Chen, and T.~Zhang.
\newblock Probabilistic bilevel coreset selection.
\newblock In \emph{{ICML}}, volume 162 of \emph{Proceedings of Machine Learning Research}, pages 27287--27302. {PMLR}, 2022.

\end{thebibliography}

\appendix
\section{Related Work}

\subsection{Sufficiency Rule}




The Sufficiency principle adopts a perspective where individuals receiving the same model decision are expected to experience similar outcomes regardless of their sensitive attributes. In this context, $A$ represents the sensitive attribute.

On the other hand, Separation deals with error rates concerning the proportion of errors relative to the ground truth. For instance, it considers the number of individuals whose loan application would have been approved but were actually denied. Sufficiency, however, considers the number of individuals who would default on their loan among those who were granted it.

From a mathematical standpoint, this distinction resembles that between recall (or true positive rate) and precision, denoted as $P [ \hat{Y} = 1|Y = 1] $ and $P [Y = 1 |\hat{Y} = 1] $, respectively.
A fairness principle that focuses on this type of error rate is known as Predictive Parity, also termed as the outcome test:

\begin{equation}
P [Y = 1 |A=a,\hat{Y} = 1]  = P [Y = 1 |A=b,\hat{Y} = 1], \forall a,b \in \mathcal{A}
\end{equation}

In other words, the model's precision should remain consistent across different sensitive groups. If we extend this requirement to apply to the scenario where $Y=0$, we obtain the following statement of conditional independence:

\begin{equation}
Y \bot A|\hat{Y}
\end{equation}

This concept is known as sufficiency.

Predictive Parity, including its broader form of sufficiency, focuses on ensuring parity in errors among individuals receiving the same decision. Predictive Parity, in particular, considers the viewpoint of the decision-maker, as they categorize individuals based on decisions rather than actual outcomes. For instance, in the context of credit lending, sufficiency is more under the control of the decision-maker than separation. This is because achieving parity given a decision is directly observable, whereas parity given truth is only known afterward. Furthermore, the group of individuals granted a loan ($\hat{Y}=1$) is less susceptible to selection bias compared to the group of loan repayments ($Y=1$). We have information only on repayment for the $\hat{Y}=1$ group, while nothing is known about the others ($\hat{Y}=0$).

Similarly, other group metrics can be defined, such as Equality of Accuracy across groups: $P [\hat{Y} = Y|A=a] = P [\hat{Y} = Y|A=b] $, for all $a,b \in \mathcal{A}$, focusing on unconditional errors, among others.




\subsubsection{Incompatibility}\label{sec:incompa}

In most cases, even in classification setting, the actual output of a model is not
a binary value, but rather a score $S \in \mathbb{R}$.


\begin{enumerate} 
\item  if $Y\not\perp A$, then it's impossible for sufficiency and independence to coexist. Consequently, if there's an imbalance in base rates among groups identified by $A$, enforcing both sufficiency and independence simultaneously is unfeasible.
\item if $Y\not\perp A$ and the distribution $(A, S, Y)$ is strictly positive, then separation and sufficiency cannot both be achieved simultaneously. This implies that separation and sufficiency can coexist only under two conditions: when there's no imbalance in sensitive groups (indicating independence of the target from sensitive attributes), or when the joint probability $(A, S, Y)$ is degenerate. In the case of binary targets, this degeneracy occurs when there are specific values of $A$ and $S$ for which only $Y=1$ (or $Y=0$) holds. In other words, separation and sufficiency coincide when the score perfectly resolves the uncertainty in the target. For instance, the ideal classifier where $S=Y$ effortlessly satisfies both sufficiency and separation.
\end{enumerate}

\subsection{Invariant Risk Minimization}

Minimizing training error leads machines into recklessly absorbing all the correlations found in training data. Understanding which patterns are useful has been previously studied as a correlation-versus-causation dilemma, since spurious correlations stemming from data biases are unrelated to the causal explanation of interest. Following this line, IRM leverage tools from causation to develop the mathematics of spurious and invariant correlations, in order to alleviate the excessive reliance of machine learning systems on data biases, allowing them to generalize to new test distributions. 

\section{Experiment Details and Results}

\subsection{Datasets}\label{sec:data_a}
\paragraph {Adult dataset}\label{adult} The Adult dataset was drawn from the 1994 United States Census Bureau data. It used personal information such as education level and working hours per week to predict whether an individual earns more or less than \$50,000 per year. The dataset is imbalanced -- the instances made less than \$50,000 constitute 25\% of the dataset, and the instances made more than \$50,000 include 75\% of the dataset. As for gender, it is also imbalanced. We use age, years of education, capital gain, capital loss, hours-per-week, etc., as continuous features, and education level, gender, etc., as categorical features.

\paragraph {COMPAS dataset}
COMPAS (Correctional Offender Management Profiling for Alternative Sanctions) is a popular commercial algorithm used by judges and parole officers for scoring criminal defendant’s likelihood of reoffending (recidivism). The COMPAS dataset includes the processed COMPAS data between 2013-2014. The data cleaning process followed the guidance in the original COMPAS repo. It Contains 6172 observations and 14 features. In our causal graph, we use 7 features. Due to the limited size of COMPAS dataset, it does not perform so well on NN based tasks. Note that we need more pre-processing on the tabular datasets. We normalize the continuous features
and use one-hot encoding to deal with the categorical features. 

\subsection{Training Details} \label{sec:training}

In our experiments, we set the following hyperparameters for optimization. In the inner loop, the model undergoes training for 100 epochs using Stochastic Gradient Descent (SGD) with a learning rate of 0.1 and momentum of 0.9. In the outer loop, the probabilities $s$ are optimized using Adam with a learning rate of 2.5 and a cosine scheduler. The outer loop is updated iteratively for 500 to 2000 times. It's worth mentioning that we employ architectures with fully connected layers for the classifier. 
\paragraph {Toxic comments}
We split the training, validation and testing set as 70\%, 10\% and 20\%. The mini-batch-size is set as 500. 

\paragraph {CelebA}
The training/validation/test set are around 82K, 18K and 18K. 
The batch-size is set as 1000. 

\paragraph {Effectiveness of Sparsity in Promoting Training Speed}\label{efficiency}
An experiment is conducted on the CelebA dataset to compare the training speed between our method with and without a sparsity constraint on sample sizes. The inclusion of the constraint significantly reduces the inner loop computation time, decreasing it from 9.24 to 5.72 hours.

\paragraph {Repetition} We repeat experiments on each dataset five times. Before each repetition, we randomly split data into training data and test data for the computation of the standard errors of the metrics.

\end{document}